\documentclass[11pt,a4paper]{article}
\usepackage[nohyperref]{acl2017}
\usepackage{times}
\usepackage{latexsym}
\usepackage{amsmath}

\usepackage{url}
\usepackage[pdftex]{graphicx} 
\aclfinalcopy

\title{Chat Detection in an Intelligent Assistant: Combining Task-oriented and Non-task-oriented Spoken Dialogue Systems}

\author{Satoshi Akasaki\thanks{\hspace{1.5mm}Work done during internship at Yahoo Japan Corporation.} \\
  The University of Tokyo \\
  {\tt akasaki@tkl.iis.u-tokyo.ac.jp} \\
  \And
  Nobuhiro Kaji \\
  Yahoo Japan Corporation\\
  {\tt nkaji@yahoo-corp.jp} \\}

\date{}

\begin{document}
\maketitle
\begin{abstract}
Recently emerged intelligent assistants on smartphones and home electronics (\textit{e.g.}, Siri and Alexa) can be seen as novel hybrids of domain-specific task-oriented spoken dialogue systems and open-domain non-task-oriented ones. To realize such hybrid dialogue systems, this paper investigates determining whether or not a user is going to have a chat with the system. To address the lack of benchmark datasets for this task, we construct a new dataset consisting of $15,160$ utterances collected from the real log data of a commercial intelligent assistant (and will release the dataset to facilitate future research activity). In addition, we investigate using tweets and Web search queries for handling open-domain user utterances, which characterize the task of chat detection. Experiments demonstrated that, while simple supervised methods are effective, the use of the tweets and search queries further improves the F$_1$-score from $86.21$ to $87.53$.
\end{abstract}

%%%%%%%%%%%%%%%%%%
\section{Introduction} \label{sec:intro}
%%%%%%%%%%%%%%%%%%

\subsection{Chat detection}
%%%%%%%%%%%%%%%%%%%%%%%

Conventional studies on spoken dialogue systems (SDS) have investigated either domain-specific task-oriented SDS%
\footnote{They can be classified as single-domain or multi-domain task-oriented SDS.} \cite{Williams07} or open-domain non-task-oriented SDS (\textit{a.k.a.}, chatbots or chat-oriented SDS) \cite{Wallace09}. The former offers convenience by helping users complete tasks in specific domains, while the latter offers entertainment through open-ended chatting (or smalltalk) with users. Although the functionalities offered by the two types of SDS are complementary to each other, little practical effort has been made to combine them. This unfortunately has limited the potential of SDS\@.

This situation is now being changed by the emergence of voice-activated intelligent assistants on smartphones and home electronics (\textit{e.g.}, Siri%
\footnote{http://www.apple.com/ios/siri} and Alexa%
\footnote{https://developer.amazon.com/alexa}). These intelligent assistants typically perform various tasks (\textit{e.g.}, Web search, weather checking, and alarm setting) while being able to have chats with users. They can be seen as a novel hybrid of multi-domain task-oriented SDS and open-domain non-task-oriented SDS\@. 

To realize such hybrid SDS, we have to determine whether or not a user is going to have a chat with the system. For example, if a user says ``\textit{What is your hobby?}'' it is considered that she is going to have a chat with the system. On the other hand, if she says ``\textit{Set an alarm at 8 o'clock},'' she is probably trying to operate her smartphone. We refer to this task as \textit{chat detection} and treat it as a binary classification problem.

Chat detection has not been explored enough in past studies. This is primarily because little attempts have been made to develop hybrids of task-oriented and non-task-oriented SDS (see Section~\ref{sec:related_work} for related work). Although task-oriented and non-task-oriented SDS have long research histories, both of them do not require chat detection. Typically, users of task-oriented SDS do not have chats with the systems and users of non-task-oriented SDS always have chats with the systems.

\subsection{Summary of this paper}
%%%%%%%%%%%%%%%%%%%%%%%%%%%%%%

In this work, we construct a new dataset for chat detection. As we already discussed, chat detection has not been explored enough, and thus there exist no benchmark datasets available. To address this situation, we collected $15,160$ user utterances from real log data of a commercial intelligent assistant, and recruited crowd workers to annotate those utterances with whether or not the users are going to have chats with the intelligent assistant. The resulting dataset will be released to facilitate future studies.

The technical challenge in chat detection is that we have to handle open-ended utterances of intelligent assistant users. Commercial intelligent assistants have a vast amount of users and they talk about a wide variety of topics especially when chatting with the assistants. It consequently becomes labor-intensive to collect a sufficiently large amount of annotated data for training accurate chat detectors.
% This situation contrasts with conventional task-oriented SDS, which handle only domain-specific utterances. 

We develop supervised binary classifiers to perform chat detection. We address the open-ended user utterances, which characterize chat detection, by using unlabeled external resources. We specifically utilize tweets (\textit{i.e.}, Twitter posts) and Web search queries to enhance the supervised classifiers.

Experimental results demonstrated that, while simple supervised methods are effective, the external resources are able to further improve them. The results demonstrated that the use of the external resources increases over $1$ point of F$_1$-score (from $86.21$ to $87.53$).

% In Section~\ref{sec:related_work} of this paper, we introduce related work. Sections~\ref{sec:data} and~\ref{sec:method} present our dataset and methods for chat detection. Section~\ref{sec:exp} presents experimental results. Section~\ref{sec:future_work} discusses future work and we conclude in Section~\ref{sec:conclusion} with a brief summary.

%%%%%%%%%%%%%%%%%%%
\section{Related Work} \label{sec:related_work}
%%%%%%%%%%%%%%%%%%%

\subsection{Previous studies on combining task-oriented and non-task-oriented SDS}
%%%%%%%%%%%%%%%%%%%%%%%%%%%%%%%%%%%%%%%%%%%%%%%%%%%%%%%%%%%%%%%%%%%%

Task-oriented and non-task-oriented SDS have long been investigated independently, and little attempts have been made to develop hybrids of the two types of SDS\@. As a consequence, previous studies have not investigated chat detection without only a few exceptions.%
\footnote{Unfortunately, we cannot discuss little about chat detection in existing commercial intelligent assistants since most of their technical details have not been disclosed. We make the best effort to compensate for it by comparing the proposed methods with our in-house intelligent assistant in the experiment.} 

Niculescu and Banchs \shortcite{Niculescu15} explored using non-task-oriented SDS as a back-off mechanism for task-oriented SDS. They, however, did not propose any concrete methods of automatically determining when to switch to non-task-oriented SDS.

\citet{Lee09} proposed an example-based dialogue manager to combine task-oriented and non-task-oriented SDS\@. In such a framework, however, it is difficult to flexibly utilize state-of-the-art supervised classifiers as a component.

Other studies proposed machine-learning-based frameworks for combining multi-domain task-oriented SDS and non-task-oriented SDS \cite{Wang14,Sarikaya17}. These assume that several components including a chat detector are already available, and explore integrating those components. They discuss little on how to develop each of the components. On the other hand, the focus of this work is to develop one of those components, a chat detector. Although it lies outside the scope of this paper to explore how to exploit chat detection method in a full dialogue system, the chat detection method is considered to serve, for example, as one component within those frameworks.

\subsection{Intent and domain determination}
%%%%%%%%%%%%%%%%%%%%%%%%%%%%%%%%%%%%%%

Chat detection is related to, but different from, intent and domain determination that have been studied in the field of SDS \cite{Guo14,Xu14,Ravuri15,Kim16,Zhang16}.

Both intent and domain determination have been investigated in domain-specific task-oriented SDS\@. Intent determination aims to determine the type of information a user is seeking in single-domain task-oriented SDS\@. For example, in the ATIS dataset, which is collected from an airline travel information service system, the information type includes flight, city, and so on \cite{Tur10}. On the other hand, domain determination aims to determine which domain is relevant to a given user utterance in multi-domain task-oriented SDS \cite{Xu14}. Note that it is possible that domain determination is followed by intent determination.

Unlike intent and domain determination, chat detection targets hybrid systems of multi-domain task-oriented SDS and open-domain non-task-oriented SDS\@, and aims to determine whether the non-task-oriented component is responsible to a given user utterance or not (\textit{i.e.}, the user is going to have a chat or not). Therefore, the objective of chat detection is different from intent and domain determination.

It may be possible to see chat detection as a specific problem of domain determination \cite{Sarikaya17}. We, nevertheless, discuss it as a different problem because of the uniqueness of the ``chat domain.'' It greatly differs from ordinary domains in that it plays a role of combining the two different types of SDS that have long been studied independently, rather than combining multiple SDS of the same types. In addition, we discuss the use of external resources, especially tweets, for chat detection. This approach is unique to chat detection and is not considered effective for ordinary domain determination.

It is interesting to note that chat detection is not followed by slot-filling unlike intent and domain determination, as far as we use a popular response generator such as seq2seq model \cite{Sutskever14} or an information retrieval based approach \cite{Yan16}. Although joint intent (or domain) determination and slot-filling has been widely studied to improve accuracy \cite{Guo14,Zhang16}, the same approach is not feasible in chat detection.

\subsection{Intelligent assistant}
%%%%%%%%%%%%%%%%%%%%%%%%%%%

Previous studies on intelligent assistants have not investigated chat detection. Their research topics are centered around those on user behaviors including the prediction of user satisfaction and engagement \cite{Jiang15,Kobayashi15,Sano16,Kiseleva16b,Kiseleva16a} and gamification \cite{Otani16}. For example, \citet{Jiang15} investigated predicting whether users are satisfied with the responses of intelligent assistants by combining diverse features including clicks and utterances. \citet{Sano16} explored predicting whether users will keep using the intelligent assistants in the future by using long-term usage histories.

Some earlier works used the Cortana dataset as a benchmark of domain determination \cite{Guo14,Xu14,Kim16} or proposed a development framework for Cortana \cite{Crook16}. Those studies, however, regarded the intelligent assistant as merely one example of multi-domain task-oriented SDS and did not explore chat detection. 

\subsection{Non-task-oriented SDS} \label{subsec:related:nontask}
%%%%%%%%%%%%%%%%%%%%%%%%%%%%%%%

Non-task-oriented SDS have long been studied in the research community. While early studies adopted rule-based methods \cite{Weizenbaum66,Wallace09}, statistical approaches have recently gained much popularity \cite{Ritter11,Vinyals15}. This research direction was pioneered by \citet{Ritter11}, who applied a phrase-based SMT model to the response generation. Later, \citet{Vinyals15} used the seq2seq model \cite{Sutskever14}. To date, a number follow-up studies have been made to improve on the response quality \cite{Hasegawa13,Shang15,Sordoni15,Li16a,Li16b,Gu16,Yan16}. Those studies assume that users always want to have chats with systems and investigate only methods of generating appropriate responses to given utterances. Chat detection is required for integrating those response generators into intelligent assistants.

\subsection{Use of conversational data}
%%%%%%%%%%%%%%%%%%%%%%%%%%%%%%%%%

The recent explosion of conversational data on the Web, especially tweets, have triggered a variety of dialogue studies. Those typically used tweets either for training response generators (\textit{c.f.}, Section~\ref{subsec:related:nontask}) or for discovering dialogue acts in an unsupervised fashion \cite{Ritter10,Higashinaka11}. This treatment of tweets differs from that in our work.

\begin{table*}[t]
\centering
\begin{tabular}{lll}
\textbf{Label} & \textbf{Example}  & \textbf{No. of votes}  \\\hline\hline
  \textsc{Chat} & Let's talk about something.  & $5$ \\  %  6457
                     &  What is your hobby?  & $7$  \\  % 1271
                     &  I don't have any holidays this month. & $5$ \\ % 8336
                     &  I'm walking around now. & $6$ \\ % 10531
                     &  Do you like cats? & $5$ \\ % 3696
                     &  You are a serious geek. & $7$ \\\hline % 11180
  \textsc{NonChat}  & Show me a picture of Mt.~Fuji.           &  $6$ \\ % 11350
                            &What's the highest building in the world?  &  $5$ \\ % 9989
                            & A nice restaurant near here. &  $7$ \\\cline{2-3} % 12860
                            & Wake me up at 9:10. & $7$ \\  % 13274
                            & Brighten the screen. & $6$ \\% 2500
                            & Turn off the alarm. & $7$ \\\hline %  3693
\end{tabular}
\caption{Example utterances and the numbers of votes. \textsc{NonChat} utterances are further divided into information seeking (top) and device control (bottom) to facilitate readers' understanding.}
\label{tab:example}
\end{table*}

%%%%%%%%%%%%%%%%%%%%%%%%%%%%
\section{Chat Detection Dataset} \label{sec:data}
%%%%%%%%%%%%%%%%%%%%%%%%%%%%

In this section we explain how we constructed the new benchmark dataset for chat detection. We then analyze the data to provide insights into the actual user behavior.

\subsection{Construction procedure} \label{subsec:data_procedure}
%%%%%%%%%%%%%%%%%%%%%%%%%%%%%%%

We sampled $15,160$ unique utterances%
\footnote{The utterances are all in Japanese. Example utterances given in this paper are English translations.} (\textit{i.e.}, automatic speech recognition results) from the real log data of a commercial intelligent assistant, Yahoo!~Voice Assist.%
\footnote{\url{https://v-assist.yahoo.co.jp}}
The log data were collected between Jan.~and Aug.~2016. In the log data, some utterances such as ``Hello'' appear frequently. To construct a dataset containing both high and low frequency utterances, we set frequency thresholds%
\footnote{We cannot disclose the exact threshold values so as to keep the detailed statistics of the original log data confidential.} to divide the utterances into three groups (high, middle, and low frequency) and then randomly sampled the same number of utterances from each of the three groups. During the data collection, we ensured privacy by manually removing utterances that included the full name of a person or detailed address information.

Next, we recruited crowd workers to annotate the $15,160$ utterances with two labels, \textsc{Chat} and \textsc{NonChat}. The workers annotated the \textsc{Chat} label when users were going to have chats with the intelligent assistant and annotated the \textsc{NonChat} label when users were seeking some information (\textit{e.g.}, searching the Web or checking the weather) or were trying to operate the smartphones (\textit{e.g.}, setting alarms or controlling volume). Note that our intelligent assistant works primarily on smartphones and thus the \textsc{NonChat} utterances include many operational instructions such as alarm setting. Example utterances are given in Table~\ref{tab:example}.

Seven workers were assigned to each utterance, and the final labels were obtained by majority vote to address the quality issue inherent in crowdsourcing. The last column in Table~\ref{tab:example} shows the number of votes that the majority label obtained. For example, five workers provided the \textsc{Chat} label (and the other two provided the \textsc{NonChat} label) to the first utterance ``Let's talk about something.''

\begin{table}[t]
\centering
\begin{tabular}{ll}
\textbf{\#Votes} & \textbf{No. of utterances}  \\ \hline
$4$ & $1701$ \\
$5$ & $2670$ \\
$6$ & $4978$ \\
$7$ & $5811$ \\
\end{tabular}
\caption{Distribution of the numbers of votes.}
\label{tab:vote}
\end{table}

\subsection{Data analysis} \label{subsec:data_analysis}
%%%%%%%%%%%%%%%%%%%%%%

The construction process described above yielded a dataset made up of $4,833$ \textsc{Chat} and $10,327$ \textsc{NonChat} utterances. 

We investigated the annotation agreement among the crowd workers. Table~\ref{tab:vote} shows the distribution of the numbers of votes that the majority labels obtained. The annotation given by the seven workers agreed perfectly in $5,811$ of the $15,160$ utterances ($38\%$). Also, at least six workers agreed in the majority of cases, $10,789$ $(=4,978+5,811)$ utterances ($71\%$). This indicates high agreement among the workers and the reliability of the annotation results.

During the data construction, we found that a typical confusing case arises when the utterance can be interpreted as an implicit information request. For example, the utterance ``\textit{I am hungry}'' can be seen as the user trying to have a chat with the assistant, but it might be the case that she is looking for a local restaurant. Similar examples include ``\textit{I have a backache}'' and so on. One solution in this case might be to ask the user a clarification question \cite{Schloder15}. Such an exploration is left for our future research.

Additionally, we manually classified the \textsc{Chat} utterances according to their dialogue acts to figure out how real users have chats with the intelligent assistant (Table~\ref{tab:dialogue_act}). The set of dialogue acts was designed by referring to \cite{Meguro10}. As shown in Table~\ref{tab:dialogue_act}, while some of the utterances are boilerplates (\textit{e.g.}, those in the \textsc{Greeting} act) and thus have limited variety, the majority of the utterances exhibit tremendous diversity. We see a wide variety of topics including private issues (\textit{e.g.}, ``\textit{I am free today}'') and questions to the assistant (\textit{e.g.}, ``\textit{Are you angry?}''). Also, we even see a movie quote (``\textit{May the force be with you}'') and a rooster crow (``\textit{Cock-a-doodle-doo}'') in the \textsc{Misc} act. These clearly represent the open-domain nature of the user utterances in intelligent assistants.

Interestingly, some users curse at the intelligent assistant probably because it failed to make appropriate responses (see the \textsc{Curse} act). Although such user behavior would not be observed from paid research participants, we observe a certain amount of curse utterances in the real data.

\begin{table}[t]
\footnotesize
\centering
\begin{tabular}{@{}l@{\;\;\;\;}l@{}}
\textbf{Dialogue act (No. of Utter.)} & \textbf{Example} \\ \hline\hline
\textsc{Greeting} ($206$) & Hello. \\
                                  &  Merry Christmas. \\ \hline
\textsc{SelfDisclosure} ($1164$) &  I am free today. \\
                                   & I have a sore throat. \\ \hline
\textsc{Order} ($716$) &Please cheer me up. \\ 
                     &          Give me a song! \\\hline 
\textsc{Question} ($1551$) & Do you have emotions? \\ 
                         & Are you angry? \\\hline 
\textsc{Invitation} ($130$) & Let's play with me! \\
                         & Let's go to karaoke next time. \\\hline 
\textsc{Information} ($214$) & My cat is acting strange. \\
                         & It snowed a lot. \\\hline 
\textsc{Thanks} ($126$) & Thank you. \\
                         & You are cool! \\\hline 
\textsc{Curse} ($172$) & You're an idiot. \\ 
                    & You are useless. \\ \hline 
\textsc{Apology} ($9$) & I'm sorry. \\ 
                    & I mistook, sorry. \\ \hline 
\textsc{Interjection} ($151$) & Whoof. \\ 
                   & Yeah, yeah.\\ \hline
\textsc{Misc} ($394$) & May the force be with you. \\ 
                    & Cock-a-doodle-doo. \\\hline
\end{tabular}
\caption{Distribution over dialogue acts and example utterances.}
\label{tab:dialogue_act}
\end{table}

%%%%%%%%%%%%%%%%%%%%%%%
\section{Detection Method} \label{sec:method}
%%%%%%%%%%%%%%%%%%%%%%%

We formulate chat detection as a binary classification problem to train supervised classifiers. In this section, we first explain the two types of classifiers explored in this paper, and then investigate the use of external resources for enhancing those classifiers.

\subsection{Base classifiers} \label{subsec:base}
%%%%%%%%%%%%%%%%%%%%%%%%

The first classifier utilizes SVM for its popularity and efficiency. It uses character and word $n$-gram ($n=1$ and $2$) features. It also uses word embedding features \cite{Turian10}. A skip-gram model \cite{Mikolov13} is trained on the entire intelligent assistant log%
\footnote{We used the same log data used in Section~\ref{sec:data}. The detailed statistics is confidential.} to learn word embeddings. The embeddings of the words in the utterance are then averaged to produce additional features.

The second classifier uses a convolutional neural network (CNN) because it has recently proven to perform well on text classification problems \cite{Kim14,Johnson15a,Johnson15b}. We follow \cite{Kim14} to develop a simple CNN that has a single convolution and max-pooling layer followed by the soft-max layer. We use a rectified linear unit (ReLU) as the non-linear activation function. The same word embeddings as SVM are used for the pre-training.

\begin{figure}[t]
\includegraphics[scale=0.17]{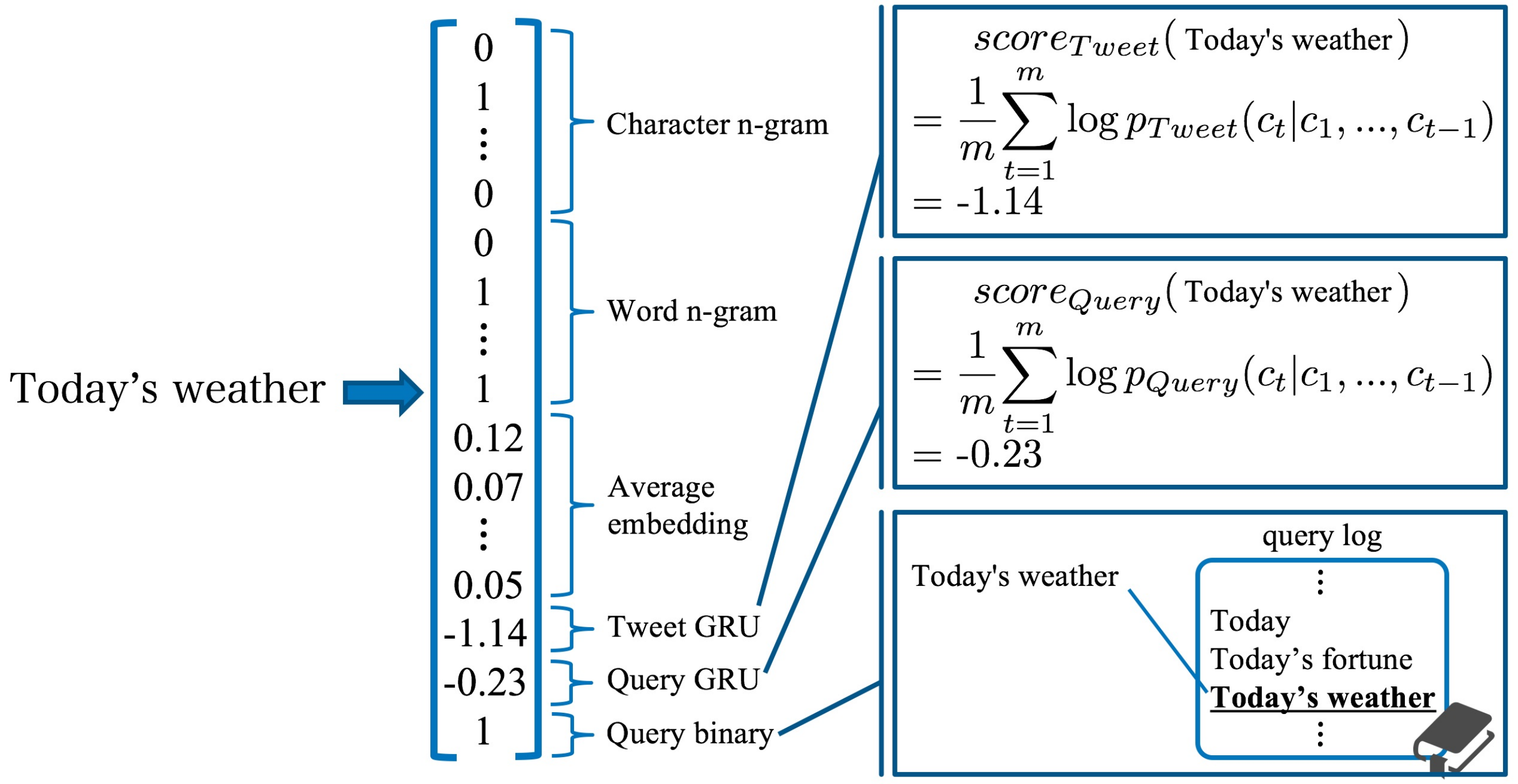} 
\caption{Feature vector representation of the example utterance ``Today's weather.'' The upper three parts of the vector represent the features described in Section~\ref{subsec:base} (character $n$-gram, word $n$-gram, and average of the word embeddings). The three additional features explained in Section~\ref{subsec:external} are added as two real-valued features (Tweet GRU and Query GRU) and one binary feature (Query binary).}
\label{fig:svm}
\end{figure}
\begin{figure}[t]
\includegraphics[scale=0.2]{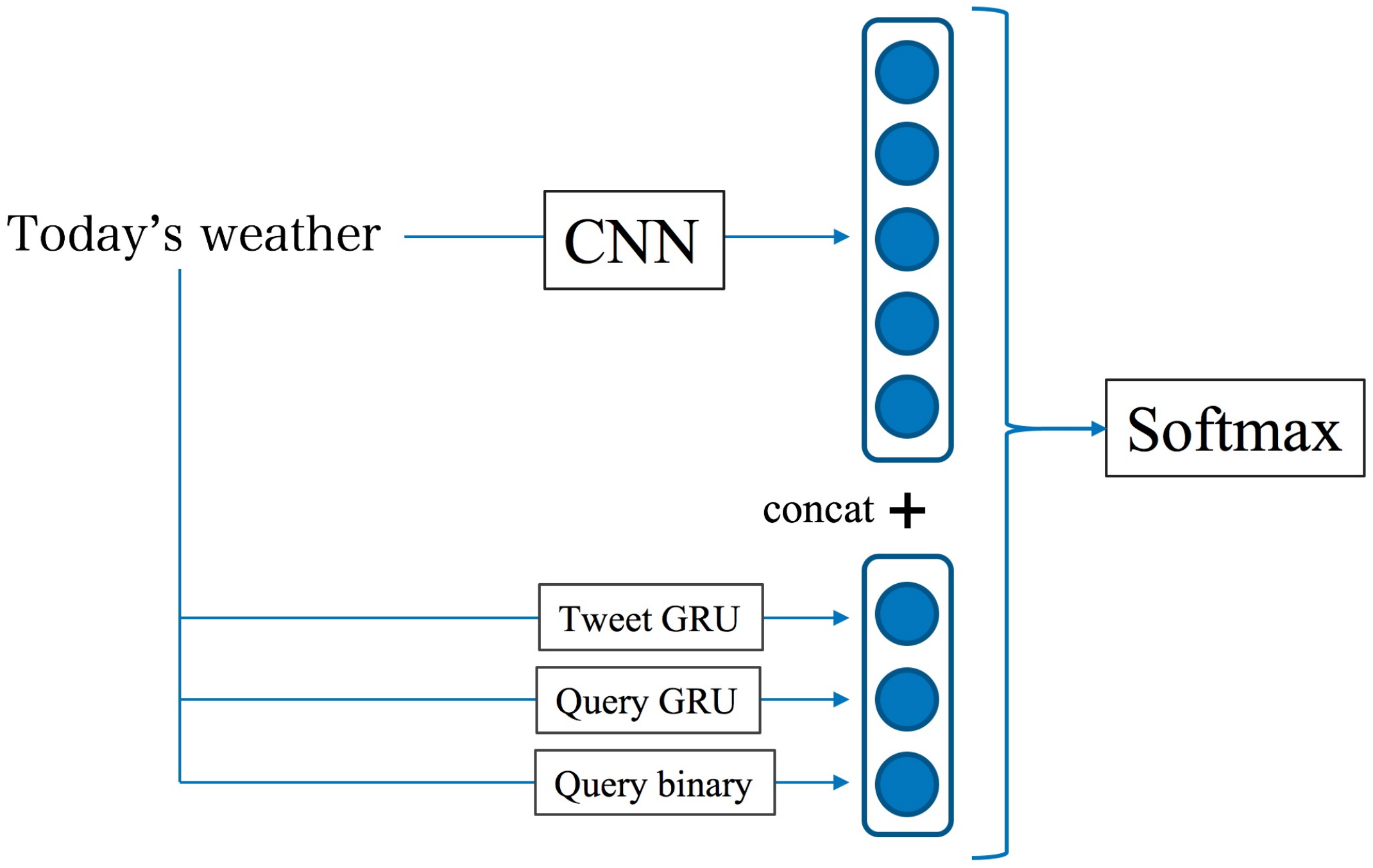} 
\caption{Architecture of our CNN-based classifier when the input utterance is ``Today's weather." The output layer of CNN and the three additional features explained in Section~\ref{subsec:external} are concatenated. The resulting vector is fed to the soft-max function.}
\label{fig:cnn}
\end{figure}

\subsection{Using external resources} \label{subsec:external}
%%%%%%%%%%%%%%%%%%%%%%%%%%%%%%%%

We next investigate using external resources for enhancing the base classifiers. Thanks to the rapid evolution of the Web in the past decade, a variety of textual data including not only conversational (\textit{i.e.}, chat-like) but also non-conversational ones are abundantly available nowadays. These data offer an effective way of enhancing the base classifiers. We specifically use tweets and Web search queries as conversational and non-conversational text data, respectively.

We train character-based%
\footnote{We also trained word-based language models in preliminary experiments and found that character-based ones perform consistently better.}
language models on tweets and Web search queries, and use their scores (\textit{i.e.}, the normalized log probabilities of the utterance) as two additional features. Let $u=c_1,c_2,\dots,c_m$ be an utterance made up of $m$ characters. Then, the score $score_{r}(u)$ of the language model trained on the external resource $r\in\{\text{tweet}, \text{query}\}$ is defined as
\begin{align*}
 score_{r}(u)=\frac{1}{m}\sum_{t=1}^{m}\log p_{r}(c_{t}\mid c_{1},\dots,c_{t-1}).
\end{align*}

The GRU language model is adopted for its superior performance \cite{Cho14,Chung14}. Let $\mathbf{x}_t$ be the embedding of $t$-th character and $\mathbf{h}_t$ be the $t$-th hidden state. GRU computes the hidden state as
\begin{align*}
\mathbf{h}_t &= (1-\mathbf{z}_t)\odot\mathbf{h}_{t-1} + \mathbf{z}_{t}\odot\tilde{\mathbf{h}}_{t} \\
\mathbf{z}_t &= \sigma(\mathbf{W}^{(z)}\mathbf{z}_t + \mathbf{U}^{(z)}\mathbf{h}_{t-1}) \\
\tilde{\mathbf{h}}_t &= \tanh(\mathbf{W}^{(h)}\mathbf{x}_t + \mathbf{U}^{(h)} (\mathbf{r}_{t}\odot\mathbf{h}_{t-1})) \\
\mathbf{r}_t &= \sigma(\mathbf{W}^{(r)}\mathbf{x}_t + \mathbf{U}^{(r)}\mathbf{h}_{t-1})
\end{align*}
where $\odot$ is the element-wise multiplication, $\sigma$ is the sigmoid and $\tanh$ is the hyperbolic tangent. $\mathbf{W}^{(z)}$, $\mathbf{U}^{(z)}$, $\mathbf{W}^{(h)}$, $\mathbf{U}^{(h)}$, $\mathbf{W}^{(r)}$, and $\mathbf{U}^{(r)}$ are weight matrices. The hidden states are fed to the soft-max to predict the next word.

We also use a binary feature indicating whether the utterance appears in the Web search query log or not. We observe that some \textsc{NonChat} utterances are made up of single entities such as location and product names. Such utterances are considered to be seeking information on those entities. We therefore use the query log as an entity dictionary to derive a feature indicating whether the utterance is likely to be a single entity.

The resulting three features are incorporated into the SVM-based classifier straightforwardly (Figure~\ref{fig:svm}). For the CNN-based classifier, they are provided as additional inputs to the soft-max layer (Figure~\ref{fig:cnn}).

%%%%%%%%%%%%%%%%%%%%%%%%%%
\section{Experimental Results} \label{sec:exp}
%%%%%%%%%%%%%%%%%%%%%%%%%%

We empirically evaluate the proposed methods on the chat detection dataset. 

\begin{table*}[t]
\centering
\begin{tabular}{lcccc}
\textbf{Model} & \textbf{Acc.} & \textbf{P} & \textbf{R} & \textbf{F$_1$} \\ \hline\hline
Majority & $68.12$ & N/A & N/A & N/A \\
Tweet GRU & $72.07$ & $54.54$ & $74.40$ & $62.94$\\
In-house IA & $78.31$ & $62.57$ & $79.51$ & $70.03$ \\  \hline
SVM                & $90.51$ & $86.42$ & $83.45$ & $84.91$ \\
SVM+embed.  & $91.35$ & $87.62$ &$84.88$ & $86.21$ \\
SVM+embed.+tweet-query & $\textbf{92.15}$ & $\textbf{88.61}$ & $\textbf{86.50}$ & $\textbf{87.53}$\\\hline
CNN                & $85.16$ & $83.40$ & $68.12$ & $74.41$ \\
CNN+pre-train.  & $90.84$ & $87.03$ & $83.80$ & $85.36$ \\
CNN+pre-train.+tweet-query & $91.48$ & $87.78$ & $85.18$ & $86.56$ \\ \hline
\end{tabular}
\caption{Chat detection results.}
\label{tab:result}
\end{table*}

\subsection{Experimental settings}
%%%%%%%%%%%%%%%%%%%%%%%%%%%%%

We performed $10$-fold cross validation on the chat detection dataset to train and evaluate the proposed classifiers. In each fold, we used $80$\%, $10$\%, and $10$\% of the data for the training, development, and evaluation, respectively.

We used \texttt{word2vec}%
\footnote{https://code.google.com/archive/p/word2vec} to learn $300$ dimensional word embeddings. They were used to induce the additional $300$ features for SVM\@. They were also used as the pre-trained word embeddings for CNN.

We used the \texttt{faster-rnn} toolkit% 
\footnote{https://github.com/yandex/faster-rnnlm} to train the GRU language models. The size of the embedding and hidden layer was set to $256$. Noise contrastive estimation \cite{Gutmann10} was used to train the soft-max function and the number of noise samples was set to $50$. Maximum entropy $4$-gram models were also trained to yield a combined model \cite{Mikolov11}.

The language models were trained on $100$ millions tweets collected between Apr.~and July 2016 and $100$ million Web search queries issued between Mar.~and Jun.~2016. The tweets were sampled from those received replies to collect only conversational tweets \cite{Ritter11}. %%The Web search queries were sampled from those appeared at least $100$ times in a day to make the data tractable size. 
The same Web search queries were used to derive the binary feature. Although it is difficult to release those data, we plan to make the feature values available together with the benchmark dataset.

We used \texttt{liblinear}%
\footnote{https://www.csie.ntu.edu.tw/\~{}cjlin/liblinear}
to train  $L_2$-regularized $L_2$-loss SVM. The hyperparameter $c$ was tuned over $\{2^{-10}, 2^{-9},\dots,2^{10}\}$.

The CNN was implemented with \texttt{chainer}.%
\footnote{http://chainer.org}
We tuned the number of feature maps over $\{100,150\}$, and filter region sizes over $\{\{2\},\{3\},\{1,2\},\{2,3\},\{3,4\},\{1,2,3\},\{2,3,4\}\}$. The mini-batch size was set to $32$. The dropout rate was set to $0.5$. We used Adam ($\alpha=0.001$, $\beta1=0.9$, $\beta2=0.999$, and $\epsilon=10^{-8}$) to perform stochastic gradient descent \cite{Kingma15}.

\subsection{Baselines}
%%%%%%%%%%%%%%%%%%%

The following baseline methods were implemented for comparison:
\begin{description}
\item[Majority] Utterances are always classified as the majority class, \textsc{NonChat}.
\item[Tweet GRU] Utterances are classified as \textsc{Chat} if the score of the GRU language model trained on the tweets exceeds a threshold. We used exactly the same GRU language model as the one that was used for deriving the feature. The threshold was calibrated on the development data by maximizing the F$_1$-score of the \textsc{Chat} class. 
\item[In-house IA] Our in-house intelligent assistant system, which adopts a hybrid of rule-based and example-based approaches. Since we cannot disclose its technical details, the result is presented just for reference.
\end{description}

\begin{table*}[t]
\centering
\begin{tabular}{llll}
\multicolumn{2}{l}{\textbf{Score (tweet/query)}} & \textbf{Label} & \textbf{Utterance} \\\hline\hline
$-0.964$ & $-1.427$ & \textsc{Chat}      & Halloween has already finished. \\
$-0.957$ & $-1.610$ & \textsc{Chat}  　　& Let's sleep. \\
$-1.233$ & $-0.562$ & \textsc{NonChat} & Pokemon Go install. \\ 
% $-0.521$ & $-2.055$ & \textsc{Chat} & Laugh. \\ 
$-1.837$ & $-0.682$ & \textsc{NonChat} & Weekly weather forecast. \\
% $-0.521$ & $-2.353$ & \textsc{NonChat} & Picture of water fall. \\
\end{tabular}
\caption{Examples of the language model scores. The first two columns represent the scores provided by the GRU language models trained on the tweets and Web search queries, respectively. The third and fourth columns represent the label and utterance.}
\label{tab:lm_example}
\end{table*}

\subsection{Result}
%%%%%%%%%%%%%%%%%

Table~\ref{tab:result} gives the precision, recall, F$_1$-score (for the \textsc{Chat} class), and overall classification accuracy results. We report only accuracy for \textbf{Majority} baseline. \textbf{+embed.}~and \textbf{+pre-train\@.}~represent using the word embedding features for SVM and the pre-trained word embeddings for CNN, respectively. \textbf{+tweet-query} represents using the three features derived from the tweets and Web search query.

Table~\ref{tab:result} represents that both of the classifiers, SVM and CNN, perform accurately. We see that both \textbf{+embed.}~and \textbf{+pre-train\@.}~improve the results. The best performing method, \textbf{SVM+embed.+tweet-query}, achieves $92$\% accuracy and $87$\% F$_1$-score, outperforming all of the baselines. CNN performed worse than SVM contrary to results reported by recent studies \cite{Kim14}. We think this is because the architecture of our CNN is rather simplistic. It might be possible to improve the CNN-based classifier by adopting more complex network, although it is likely to come at the cost of extra training time. Another reason would be that our SVM classifier uses carefully designed features beyond word $1$-grams.

\begin{table}[t]
\centering
\begin{tabular}{lllll}
\textbf{Feature} & \textbf{Acc.} & \textbf{P} & \textbf{R} & \textbf{F$_1$} \\ \hline
tweet GRU & $91.53$ & $87.62$ & $85.49$ & $86.53$ \\
query GRU &  $91.38$ & $87.55$ & $85.06$ & $86.28$ \\
query binary & $91.42$ & $87.56$ & $85.21$ & $86.36$ \\
%All & $\textbf{92.15}$ & $\textbf{88.61}$ & $\textbf{86.50}$ & $\textbf{87.53}$\\\hline
\end{tabular}
\caption{Effect of the three features derived from the tweets and Web search queries.}
\label{tab:result_detail}
\end{table}

Table~\ref{tab:result} also represents that the external resources are effective, improving F$_1$-scores almost $1$ points in both SVM and CNN. Table~\ref{tab:lm_example} illustrates example utterances and their language model scores. We see that the language models trained on the tweets and queries successfully provide the \textsc{Chat} utterances with high and low scores, respectively. Table~\ref{tab:result_detail} shows chat detection results when each of the three features derived from the external resources is added to \textbf{SVM+embed.} The results represent that they are all worse than \textbf{SVM+embed.+tweet-query} and thus it is crucial to combine all of them for achieving the best performance.

\begin{table}[t]
\centering
\begin{tabular}{cc|cc}
\textbf{Feature} & \textbf{Weight} & \textbf{Feature} & \textbf{Weight} \\ \hline
tweet GRU & $1.128$ & query GRU & $-0.771$ \\
%embed_{266}        & $0.250$ & embed_{208}        & $-0.278$ \\
I                & $0.215$ & call to       & $-0.217$ \\
Sing           & $0.195$ & volume     & $-0.196$ \\ 
\end{tabular}
\caption{Examples feature weights of \textbf{SVM+embed+tweet-query}.}
\label{tab:feature}
\end{table}

\begin{table}[t]
\centering
\begin{tabular}{@{}l@{\;\;\;}lllll@{}}
\textbf{\#Votes} & \textbf{\#Utter.} & \textbf{Acc.} & \textbf{P} & \textbf{R} & \textbf{F$_1$} \\ \hline
$4$ & $1701$ & $66.67$ & $55.41$ & $59.81$ & $57.53$ \\
$5$ & $2670$ & $87.72$ & $80.46$ & $83.01$ & $81.72$ \\
$6$ & $4978$ & $96.02$ & $92.73$ & $93.87$ & $93.30$ \\
$7$ & $5811$ & $98.33$ & $96.73$ & $97.68$ & $97.20$ \\
\end{tabular}
\caption{Chat detection results across the numbers of votes that the majority label obtained.}
\label{tab:result_per_vote}
\end{table}

Table~\ref{tab:feature} shows examples of feature weights of \textbf{SVM+embed.+tweet-query}. Tweet GRU and query GRU denote the language model score features. The others are word $n$-gram features. We see that the language model scores have the large positive and negative weights, respectively. This indicates that effectiveness of the language models. We also see that the first person has a large positive weight, while terms related to device controlling (``call to'' and ``volume'') have large negative weights.

Table~\ref{tab:result_per_vote} represents chat detection results of \textbf{SVM+embd.+tweet-query} across the numbers of votes that the majority label obtained. As expected, we see that all metrics get higher as the number of agreement among the crowd workers becomes larger. In fact, we see as much as $98$\% accuracy when all seven workers agree. This implies that utterances easy for humans to classify are also easy for the classifiers.

\subsection{Training data size}
%%%%%%%%%%%%%%%%%%%%%%%%%%

We next investigate the effect of the training data size on the classification accuracy. 

\begin{figure}[t]
\centering
\includegraphics[width=0.45\textwidth]{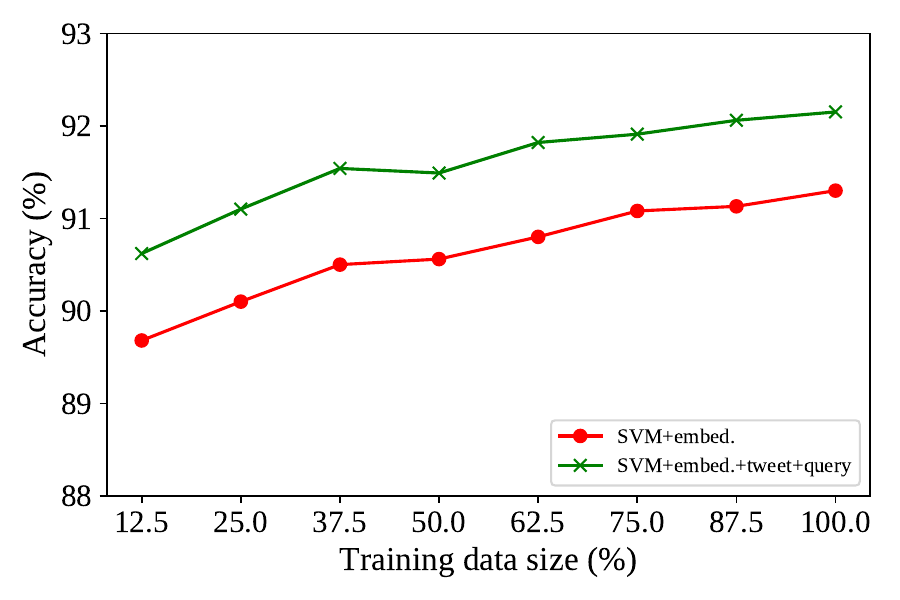} 
\caption{Learning curve of the proposed methods. The horizontal axis represents what percentage of the training portion is used in each fold of the cross validation. The vertical axis represents the classification accuracy.}
\label{fig:learning_curve}
\end{figure}

Figure~\ref{fig:learning_curve} illustrates the learning curve. It represents that the classification accuracy improves almost monotonically as the training data size increases. Although our training data is by no means small, the shape of the learning curve nevertheless suggests that further improvement would be achieved by adding more training data. This implies that a very large amount of training data are required for covering open-domain utterances in intelligent assistants. 

The figure at the same time represents the usefulness of the external resources. We see that \textbf{SVM+embed.+tweet-query} trained on about $25$\% of the training data is able to achieve comparable accuracy with \textbf{SVM+embed.}~trained on the entire training data. This result suggests that the external resources are able to compensate for the scarcity of annotated data.

\subsection{Utterance length}
%%%%%%%%%%%%%%%%%%%%%%%%%

\begin{figure}[t]
\centering
\includegraphics[width=0.45\textwidth]{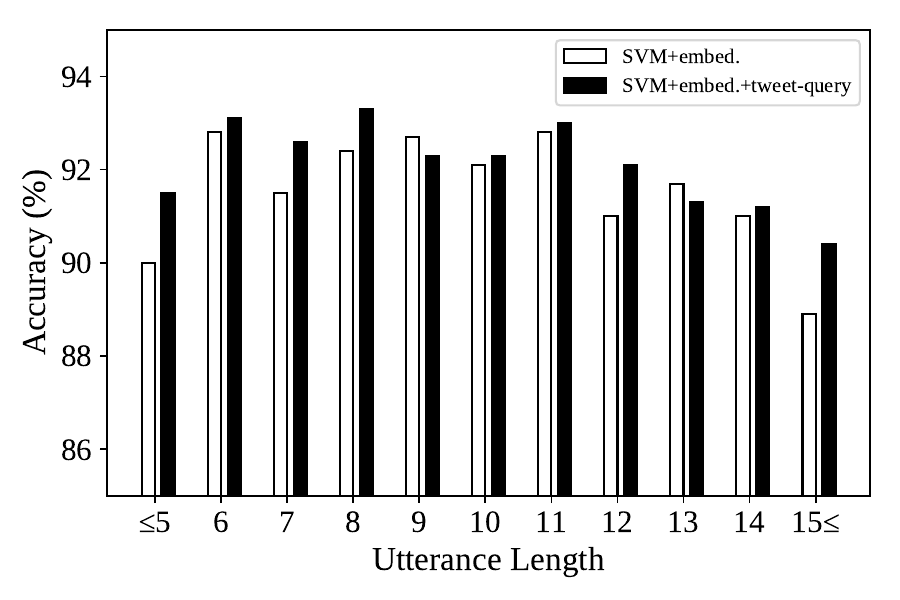} 
\caption{Classification accuracy across utterance lengths in the number of characters.}
\label{fig:length}
\end{figure}

We finally investigate how the utterance length correlates with the classification accuracy. Figure~\ref{fig:length} illustrates the classification accuracies of \textbf{SVM+embed.}~and \textbf{SVM+embed.+tweet-query} for each utterance length in the number of characters.

Figure~\ref{fig:length} reveals that the difference between the two proposed methods is evident in short utterances (\textit{i.e.}, $\leq 5$). This is because those utterances are too short to contain sufficient information required for classification, and the additional features are helpful. We note that Japanese writing system uses ideograms and thus even five characters is enough to represent a simple sentence.

We also see a clear difference in longer utterances (\textit{i.e.}, $15\leq$) as well. We consider those long utterances are difficult to classify because some words in the utterances are irrelevant for the classification and the $n$-gram and embedding features include those irrelevant ones. On the other hand, we consider that the language model scores are good at capturing stylistic information irrespective of the utterance length.

%%%%%%%%%%%%%%%%%%
\section{Future Work} \label{sec:future_work}
%%%%%%%%%%%%%%%%%%

As discussed in Section~\ref{subsec:data_analysis}, some user utterances such as ``\textit{I am hungry}'' are ambiguous in nature and thus are difficult to handle in the current framework. An important future work is to develop a sophisticated dialogue manager to handle such utterances, for example, by making clarification questions \cite{Schloder15}.

We manually investigated the dialogue acts in the chat detection dataset (\textit{c.f.}, Section~\ref{subsec:data_analysis}). It is interesting to automatically determine the dialogue acts to help producing appropriate system responses. Some related studies exist in such a research direction \cite{Meguro10}.

Although we used only text data to perform chat detection, we can also utilize contextual information such as the previous utterances \cite{Xu14}, the acoustic information \cite{Jiang15}, and the user profile \cite{Sano16}. It is an interesting research topic to use such contextual information beyond text. It is considered promising to make use of a neural network for integrating such heterogeneous information.

An automatic speech recognition (ASR) error is a popular problem in SDS, and previous studies have proposed sophisticated techniques, including re-ranking \cite{Morbini12} and POMDP \cite{Williams07}, for addressing the ASR errors. Incorporating these techniques into our methods is also an important future work.

Although the studies on non-task-oriented SDS have made substantial progress in the past few years, it unfortunately remains difficult for the systems to fluently chat with users \cite{Higashinaka15}. Further efforts on improving non-task-oriented dialogue systems is an important future work.

%%%%%%%%%%%%%%%%%
\section{Conclusion} \label{sec:conclusion}
%%%%%%%%%%%%%%%%%

This paper investigated chat detection for combining domain-specific task-oriented SDS and open-domain non-task-oriented SDS\@. To address the scarcity of benchmark datasets for this task, we constructed a new benchmark dataset from the real log data of a commercial intelligent assistant. In addition, we investigated using the external resources, tweets and Web search queries, to handle open-domain user utterances, which characterize the task of chat detection. The empirical experiment demonstrated that the off-the-shelf supervised methods augmented with the external resources perform accurately, outperforming the baseline approaches. We hope that this study contributes to remove the long-standing boundary between task-oriented and non-task-oriented SDS\@.

To facilitate future research, we are going to release the dataset together with the feature values derived from the tweets and Web search queries.%
\footnote{\url{https://research-lab.yahoo.co.jp/en/software}}

%%%%%%%%%%%%%%%%%%%%%%%%
\section*{Acknowledgments}
%%%%%%%%%%%%%%%%%%%%%%%%

We thank Manabu Sassano, Chikara Hashimoto, Naoki Yoshinaga, and Masashi Toyoda for fruitful discussions and comments. We also thank the anonymous reviewers.

% at Yahoo Japan Corporation, and Naoki Yoshinaga at the University of Tokyo

\bibliographystyle{acl_natbib}
\bibliography{acl2017}  

\begin{thebibliography}{}
\expandafter\ifx\csname natexlab\endcsname\relax\def\natexlab#1{#1}\fi

\bibitem[{Cho et~al.(2014)Cho, van Merrienboer, Gulcehre, Bahdanau, Bougares,
  Schwenk, and Bengio}]{Cho14}
Kyunghyun Cho, Bart van Merrienboer, Caglar Gulcehre, Dzmitry Bahdanau, Fethi
  Bougares, Holger Schwenk, and Yoshua Bengio. 2014.
\newblock \href{http://www.aclweb.org/anthology/D14-1179}{Learning phrase
  representations using rnn encoder--decoder for statistical machine
  translation}.
\newblock In {\em Proceedings of EMNLP\/}. pages 1724--1734.
\newblock
  \href{http://www.aclweb.org/anthology/D14-1179}{http://www.aclweb.org/anthology/D14-1179}.

\bibitem[{Chung et~al.(2014)Chung, Gulcehre, Cho, and Bengio}]{Chung14}
Junyoung Chung, Caglar Gulcehre, KyungHyun Cho, and Yoshua Bengio. 2014.
\newblock Empirical evaluation of gated recurrent neural networks on sequence
  modeling.
\newblock arXiv:1412.3555.

\bibitem[{Crook et~al.(2016)Crook, Marin, Agarwal, Aggarwal, Anastasakos,
  Bikkula, Boies, Celikyilmaz, Chandramohan, Feizollahi, Holenstein, Jeong,
  Khan, Kim, Krawczyk, Liu, Panic, Radostev, Ramesh, Robichaud, Rochette,
  Stromberg, and Sarikaya}]{Crook16}
Paul Crook, Alex Marin, Vipul Agarwal, Khushboo Aggarwal, Tasos Anastasakos,
  Ravi Bikkula, Daniel Boies, Asli Celikyilmaz, Senthilkumar Chandramohan,
  Zhaleh Feizollahi, Roman Holenstein, Minwoo Jeong, Omar Khan, Young-Bum Kim,
  Elizabeth Krawczyk, Xiaohu Liu, Danko Panic, Vasiliy Radostev, Nikhil Ramesh,
  Jean-Phillipe Robichaud, Alexandre Rochette, Logan Stromberg, and Ruhi
  Sarikaya. 2016.
\newblock \href{http://www.aclweb.org/anthology/N16-3010}{Task completion
  platform: A self-serve multi-domain goal oriented dialogue platform}.
\newblock In {\em Proceedings of NAACL (Demonstrations)\/}. pages 47--51.
\newblock
  \href{http://www.aclweb.org/anthology/N16-3010}{http://www.aclweb.org/anthology/N16-3010}.

\bibitem[{Gu et~al.(2016)Gu, Lu, Li, and Li}]{Gu16}
Jiatao Gu, Zhengdong Lu, Hang Li, and Victor~O.K. Li. 2016.
\newblock \href{http://www.aclweb.org/anthology/P16-1154}{Incorporating copying
  mechanism in sequence-to-sequence learning}.
\newblock In {\em Proceedings of ACL\/}. pages 1631--1640.
\newblock
  \href{http://www.aclweb.org/anthology/P16-1154}{http://www.aclweb.org/anthology/P16-1154}.

\bibitem[{Guo et~al.(2014)Guo, Tur, tau Yih, and Zweig}]{Guo14}
Daniel~(Zhaohan) Guo, Gokhan Tur, Scott~Wen tau Yih, and Geoffrey Zweig. 2014.
\newblock Joint semantic utterance classification and slot filling with
  recursive neural networks.
\newblock In {\em Proceedings of IEEE SLT Workshop\/}.

\bibitem[{Gutmann and Hyv\"{a}rinen(2010)}]{Gutmann10}
Michael Gutmann and Aapo Hyv\"{a}rinen. 2010.
\newblock Noise-contrastive estimation: A new estimation principle for
  unnormalized statistical models.
\newblock In {\em Proceedings of AISTATS\/}. pages 297--304.

\bibitem[{Hasegawa et~al.(2013)Hasegawa, Kaji, Yoshinaga, and
  Toyoda}]{Hasegawa13}
Takayuki Hasegawa, Nobuhiro Kaji, Naoki Yoshinaga, and Masashi Toyoda. 2013.
\newblock \href{http://www.aclweb.org/anthology/P13-1095}{Predicting and
  eliciting addressee's emotion in online dialogue}.
\newblock In {\em Proceedings of ACL\/}. pages 964--972.
\newblock
  \href{http://www.aclweb.org/anthology/P13-1095}{http://www.aclweb.org/anthology/P13-1095}.

\bibitem[{Higashinaka et~al.(2015)Higashinaka, Funakoshi, Araki, Tsukahara,
  Kobayashi, and Mizukami}]{Higashinaka15}
Ryuichiro Higashinaka, Kotaro Funakoshi, Masahiro Araki, Hiroshi Tsukahara,
  Yuka Kobayashi, and Masahiro Mizukami. 2015.
\newblock \href{http://aclweb.org/anthology/W15-4611}{Towards taxonomy of
  errors in chat-oriented dialogue systems}.
\newblock In {\em Proceedings of SIGDIAL\/}. pages 87--95.
\newblock
  \href{http://aclweb.org/anthology/W15-4611}{http://aclweb.org/anthology/W15-4611}.

\bibitem[{Higashinaka et~al.(2011)Higashinaka, Kawamae, Sadamitsu, Minami,
  Meguro, Dohsaka, and Inagaki}]{Higashinaka11}
Ryuichiro Higashinaka, Noriaki Kawamae, Kugatsu Sadamitsu, Yasuhiro Minami,
  Toyomi Meguro, Kohji Dohsaka, and Hirohito Inagaki. 2011.
\newblock Building a conversational model from two-tweets.
\newblock In {\em Proceedings of ASRU\/}. pages 330--335.

\bibitem[{Jiang et~al.(2015)Jiang, Awadallah, Jones, Ozertem, Zitouni,
  Kulkarni, and Khan}]{Jiang15}
Jiepu Jiang, {Ahmed Hassan} Awadallah, Rosie Jones, Umut Ozertem, Imed Zitouni,
  {Ranjitha Gurunath} Kulkarni, and Omar~Zia Khan. 2015.
\newblock Automatic online evaluation of intelligent assistants.
\newblock In {\em Proceedings of WWW\/}. pages 506--516.

\bibitem[{Johnson and Zhang(2015{\natexlab{a}})}]{Johnson15a}
Rie Johnson and Tong Zhang. 2015{\natexlab{a}}.
\newblock \href{http://www.aclweb.org/anthology/N15-1011}{Effective use of word
  order for text categorization with convolutional neural networks}.
\newblock In {\em Proceedings of NAACL\/}. pages 103--112.
\newblock
  \href{http://www.aclweb.org/anthology/N15-1011}{http://www.aclweb.org/anthology/N15-1011}.

\bibitem[{Johnson and Zhang(2015{\natexlab{b}})}]{Johnson15b}
Rie Johnson and Tong Zhang. 2015{\natexlab{b}}.
\newblock Semi-supervised convolutional neural networks for text categorization
  via region embedding.
\newblock In {\em Advances in NIPS\/}, pages 919--927.

\bibitem[{Kim et~al.(2016)Kim, Tur, Celikyilmaz, Cao, and Wang}]{Kim16}
Joo-Kyung Kim, Gokhan Tur, Asli Celikyilmaz, Bin Cao, and Ye-Yi Wang. 2016.
\newblock Intent detection using semantically enriched word embeddings.
\newblock In {\em Proceedings of IEEE SLT Workshop\/}.

\bibitem[{Kim(2014)}]{Kim14}
Yoon Kim. 2014.
\newblock \href{http://www.aclweb.org/anthology/D14-1181}{Convolutional neural
  networks for sentence classification}.
\newblock In {\em Proceedings of EMNLP\/}. pages 1746--1751.
\newblock
  \href{http://www.aclweb.org/anthology/D14-1181}{http://www.aclweb.org/anthology/D14-1181}.

\bibitem[{Kingma and Ba(2015)}]{Kingma15}
Diederik~P. Kingma and Jimmy Ba. 2015.
\newblock Adam: A method for stochastic optimization.
\newblock In {\em Proceedings of ICLR\/}.

\bibitem[{Kiseleva et~al.(2016{\natexlab{a}})Kiseleva, Williams, Awadallah,
  Crook, Zitouni, and Anastasakos}]{Kiseleva16b}
Julia Kiseleva, Kyle Williams, {Ahmed Hassan} Awadallah, Aidan Crook, Imed
  Zitouni, and Tasos Anastasakos. 2016{\natexlab{a}}.
\newblock Predicting user satisfaction with intelligent assistants.
\newblock In {\em Proceedings of SIGIR\/}. pages 45--54.

\bibitem[{Kiseleva et~al.(2016{\natexlab{b}})Kiseleva, Williams, Awadallah,
  Crook, Zitouni, and Anastasakos}]{Kiseleva16a}
Julia Kiseleva, Kyle Williams, Ahmed~Hassan Awadallah, Aidan~C. Crook, Imed
  Zitouni, and Tasos Anastasakos. 2016{\natexlab{b}}.
\newblock Understanding user satisfaction with intelligent assistants.
\newblock In {\em Proceedings of SIGCHIIR\/}. pages 121--130.

\bibitem[{Kobayashi et~al.(2015)Kobayashi, Tanio, and Sassano}]{Kobayashi15}
Hayato Kobayashi, Kaori Tanio, and Manabu Sassano. 2015.
\newblock \href{http://aclweb.org/anthology/W15-4656}{Effects of game on user
  engagement with spoken dialogue system}.
\newblock In {\em Proceedings of SIGDIAL\/}. pages 422--426.
\newblock
  \href{http://aclweb.org/anthology/W15-4656}{http://aclweb.org/anthology/W15-4656}.

\bibitem[{Lee et~al.(2007)Lee, Jung, Kim, and Lee}]{Lee09}
Cheongjae Lee, Sangkeun Jung, Seokhwan Kim, and Gary~Geunbae Lee. 2007.
\newblock Example-based dialog modeling for practical multi-domain dialog
  system.
\newblock {\em Speech Communication\/} 51(5):466--484.

\bibitem[{Li et~al.(2016{\natexlab{a}})Li, Galley, Brockett, Gao, and
  Dolan}]{Li16a}
Jiwei Li, Michel Galley, Chris Brockett, Jianfeng Gao, and Bill Dolan.
  2016{\natexlab{a}}.
\newblock \href{http://www.aclweb.org/anthology/N16-1014}{A diversity-promoting
  objective function for neural conversation models}.
\newblock In {\em Proceedings of NAACL\/}. pages 110--119.
\newblock
  \href{http://www.aclweb.org/anthology/N16-1014}{http://www.aclweb.org/anthology/N16-1014}.

\bibitem[{Li et~al.(2016{\natexlab{b}})Li, Galley, Brockett, Spithourakis, Gao,
  and Dolan}]{Li16b}
Jiwei Li, Michel Galley, Chris Brockett, Georgios Spithourakis, Jianfeng Gao,
  and Bill Dolan. 2016{\natexlab{b}}.
\newblock \href{http://www.aclweb.org/anthology/P16-1094}{A persona-based
  neural conversation model}.
\newblock In {\em Proceedings of ACL\/}. pages 994--1003.
\newblock
  \href{http://www.aclweb.org/anthology/P16-1094}{http://www.aclweb.org/anthology/P16-1094}.

\bibitem[{Meguro et~al.(2010)Meguro, Higashinaka, Minami, and
  Dohsaka}]{Meguro10}
Toyomi Meguro, Ryuichiro Higashinaka, Yasuhiro Minami, and Kohji Dohsaka. 2010.
\newblock \href{http://www.aclweb.org/anthology/C10-1086}{Controlling
  listening-oriented dialogue using partially observable markov decision
  processes}.
\newblock In {\em Proceedings of Coling\/}. pages 761--769.
\newblock
  \href{http://www.aclweb.org/anthology/C10-1086}{http://www.aclweb.org/anthology/C10-1086}.

\bibitem[{Mikolov et~al.(2011)Mikolov, Deoras, Povey, Burget, and
  Cernocky}]{Mikolov11}
Tomas Mikolov, Anoop Deoras, Daniel Povey, Lukas Burget, and Jan Cernocky.
  2011.
\newblock Strategies for training large scale neural network language models.
\newblock In {\em Proceedings of ASRU\/}. pages 196–--201.

\bibitem[{Mikolov et~al.(2013)Mikolov, Sutskever, Chen, Corrado, and
  Dean}]{Mikolov13}
Tomas Mikolov, Ilya Sutskever, Kai Chen, Greg Corrado, and Jeffrey Dean. 2013.
\newblock Distributed representations of words and phrases and their
  compositionality.
\newblock In {\em Advances in NIPS\/}. pages 3111--3119.

\bibitem[{Morbini et~al.(2012)Morbini, Audhkhasi, Artstein, {Van Segbroeck},
  Sagae, Georgiou, {R. Traum}, and Narayanan}]{Morbini12}
Fabrizio Morbini, Kartik Audhkhasi, Ron Artstein, Maarten {Van Segbroeck},
  Kenji Sagae, Panayiotis Georgiou, David {R. Traum}, and Shri Narayanan. 2012.
\newblock A reranking approach for recognition and classification of speech
  input in conversational dialogue systems.
\newblock In {\em Proceedings of SLT\/}. pages 49--54.

\bibitem[{Niculescu and Banchs(2015)}]{Niculescu15}
Andreea~I. Niculescu and Rafael~E. Banchs. 2015.
\newblock Strategies to cope with errors in human-machine speech interactions:
  using chatbots as back-off mechanism for task-oriented dialogues.
\newblock In {\em Proceedings of ERRARE\/}.

\bibitem[{Otani et~al.(2016)Otani, Kawahara, Kurohashi, Kaji, and
  Sassano}]{Otani16}
Naoki Otani, Daisuke Kawahara, Sadao Kurohashi, Nobuhiro Kaji, and Manabu
  Sassano. 2016.
\newblock \href{http://aclweb.org/anthology/W16-4402}{Large-scale acquisition
  of commonsense knowledge via a quiz game on a dialogue system}.
\newblock In {\em Proceedings of OKBQA\/}. pages 11--20.
\newblock
  \href{http://aclweb.org/anthology/W16-4402}{http://aclweb.org/anthology/W16-4402}.

\bibitem[{Ravuri and Stolcke(2015)}]{Ravuri15}
Suman Ravuri and Andreas Stolcke. 2015.
\newblock A comparative study of neural network models for lexical intent
  classification.
\newblock In {\em In Proceedings of ASRU\/}. pages 368--374.

\bibitem[{Ritter et~al.(2010)Ritter, Cherry, and Dolan}]{Ritter10}
Alan Ritter, Colin Cherry, and Bill Dolan. 2010.
\newblock \href{http://www.aclweb.org/anthology/N10-1020}{Unsupervised modeling
  of twitter conversations}.
\newblock In {\em In Proceedings of NAACL\/}. pages 172--180.
\newblock
  \href{http://www.aclweb.org/anthology/N10-1020}{http://www.aclweb.org/anthology/N10-1020}.

\bibitem[{Ritter et~al.(2011)Ritter, Cherry, and Dolan}]{Ritter11}
Alan Ritter, Colin Cherry, and William~B. Dolan. 2011.
\newblock \href{http://www.aclweb.org/anthology/D11-1054}{Data-driven response
  generation in social media}.
\newblock In {\em Proceedings of EMNLP\/}. pages 583--593.
\newblock
  \href{http://www.aclweb.org/anthology/D11-1054}{http://www.aclweb.org/anthology/D11-1054}.

\bibitem[{Sano et~al.(2016)Sano, Kaji, and Sassano}]{Sano16}
Shumpei Sano, Nobuhiro Kaji, and Manabu Sassano. 2016.
\newblock \href{http://www.aclweb.org/anthology/P16-1114}{Prediction of
  prospective user engagement with intelligent assistants}.
\newblock In {\em Proceedings of ACL\/}. pages 1203--1212.
\newblock
  \href{http://www.aclweb.org/anthology/P16-1114}{http://www.aclweb.org/anthology/P16-1114}.

\bibitem[{Sarikaya(2017)}]{Sarikaya17}
Ruhi Sarikaya. 2017.
\newblock The technology behind personal digital assistants: An overview of the
  system architecture and key components.
\newblock {\em IEEE Signal Processing Magazine\/} 34(1):67--81.

\bibitem[{Schl\"{o}der and Fernandez(2015)}]{Schloder15}
Julian~J. Schl\"{o}der and Raquel Fernandez. 2015.
\newblock \href{http://www.aclweb.org/anthology/W15-0106}{Clarifying intentions
  in dialogue: A corpus study}.
\newblock In {\em Proceedings of the 11th International Conference on
  Computational Semantics\/}. pages 46--51.
\newblock
  \href{http://www.aclweb.org/anthology/W15-0106}{http://www.aclweb.org/anthology/W15-0106}.

\bibitem[{Shang et~al.(2015)Shang, Lu, and Li}]{Shang15}
Lifeng Shang, Zhengdong Lu, and Hang Li. 2015.
\newblock \href{http://www.aclweb.org/anthology/P15-1152}{Neural responding
  machine for short-text conversation}.
\newblock In {\em Proceedings of ACL\/}. pages 1577--1586.
\newblock
  \href{http://www.aclweb.org/anthology/P15-1152}{http://www.aclweb.org/anthology/P15-1152}.

\bibitem[{Sordoni et~al.(2015)Sordoni, Galley, Auli, Brockett, Ji, Mitchell,
  Nie, Gao, and Dolan}]{Sordoni15}
Alessandro Sordoni, Michel Galley, Michael Auli, Chris Brockett, Yangfeng Ji,
  Margaret Mitchell, Jian-Yun Nie, Jianfeng Gao, and Bill Dolan. 2015.
\newblock \href{http://www.aclweb.org/anthology/N15-1020}{A neural network
  approach to context-sensitive generation of conversational responses}.
\newblock In {\em Proceedings of NAACL\/}. pages 196--205.
\newblock
  \href{http://www.aclweb.org/anthology/N15-1020}{http://www.aclweb.org/anthology/N15-1020}.

\bibitem[{Sutskever et~al.(2014)Sutskever, Vinyals, and Le}]{Sutskever14}
Ilya Sutskever, Oriol Vinyals, and Quoc~V Le. 2014.
\newblock Sequence to sequence learning with neural networks.
\newblock In {\em Advances in NIPS\/}, pages 3104--3112.

\bibitem[{Tur et~al.(2010)Tur, Hakkani-T{\"u}r, and Heck}]{Tur10}
Gokhan Tur, Dilek Hakkani-T{\"u}r, and Larry Heck. 2010.
\newblock What is left to be understood in atis?
\newblock In {\em Proceedings of IEEE SLT Workshop\/}. pages 19--24.

\bibitem[{Turian et~al.(2010)Turian, Ratinov, and Bengio}]{Turian10}
Joseph Turian, Lev-Arie Ratinov, and Yoshua Bengio. 2010.
\newblock \href{http://www.aclweb.org/anthology/P10-1040}{Word representations:
  A simple and general method for semi-supervised learning}.
\newblock In {\em Proceedings of ACL\/}. pages 384--394.
\newblock
  \href{http://www.aclweb.org/anthology/P10-1040}{http://www.aclweb.org/anthology/P10-1040}.

\bibitem[{Vinyals and Le(2015)}]{Vinyals15}
Oriol Vinyals and Quoc Le. 2015.
\newblock A neural conversational model.
\newblock In {\em Proceedings of Deep Learning Workshop\/}.

\bibitem[{Wallace(2009)}]{Wallace09}
{Richard S.} Wallace. 2009.
\newblock {\em The Anatomy of A.L.I.C.E.\/}, Springer, pages 181--210.

\bibitem[{Wang et~al.(2014)Wang, Chen, Wang, Tian, Wu, and Wang}]{Wang14}
Zhuoran Wang, Hongliang Chen, Guanchun Wang, Hao Tian, Hua Wu, and Haifeng
  Wang. 2014.
\newblock \href{http://www.aclweb.org/anthology/D14-1007}{Policy learning for
  domain selection in an extensible multi-domain spoken dialogue system}.
\newblock In {\em Proceedings of EMNLP\/}. pages 57--67.
\newblock
  \href{http://www.aclweb.org/anthology/D14-1007}{http://www.aclweb.org/anthology/D14-1007}.

\bibitem[{Weizenbaum(1966)}]{Weizenbaum66}
Joseph Weizenbaum. 1966.
\newblock Eliza--a computer program for the study of natural language
  communication between man and machine.
\newblock {\em Communications of the ACM\/} 9(1):36--45.

\bibitem[{Williams and Young(2007)}]{Williams07}
{Jason D.} Williams and Steve Young. 2007.
\newblock Partially observable markov decision processes for spoken dialog
  systems.
\newblock {\em Computer Speech \& Language\/} 21(2):393--422.

\bibitem[{Xu and Sarikaya(2014)}]{Xu14}
Puyang Xu and Ruhi Sarikaya. 2014.
\newblock Contextual domain classification in spoken language understanding
  systems using recurrent neural network.
\newblock In {\em Proceedings of ICASSP\/}. pages 136--140.

\bibitem[{Yan et~al.(2016)Yan, Duan, Bao, Chen, Zhou, Li, and Zhou}]{Yan16}
Zhao Yan, Nan Duan, Junwei Bao, Peng Chen, Ming Zhou, Zhoujun Li, and Jianshe
  Zhou. 2016.
\newblock \href{http://www.aclweb.org/anthology/P16-1049}{Docchat: An
  information retrieval approach for chatbot engines using unstructured
  documents}.
\newblock In {\em Proceedings of ACL\/}. pages 516--525.
\newblock
  \href{http://www.aclweb.org/anthology/P16-1049}{http://www.aclweb.org/anthology/P16-1049}.

\bibitem[{Zhang and Wang(2016)}]{Zhang16}
Xiaodong Zhang and Houfeng Wang. 2016.
\newblock A joint model of intent determination and slot filling for spoken
  language understanding.
\newblock In {\em Proceedings of IJCAI\/}. pages 2993--2999.

\end{thebibliography}
\end{document}